\pdfoutput=1

\documentclass[11pt]{article}

\usepackage[final]{acl}
\usepackage{url}
\usepackage{times}
\usepackage{multirow}
\usepackage{enumitem}
\usepackage{latexsym}
\usepackage{xcolor}
\usepackage{array}
\usepackage{geometry}
\usepackage{booktabs}
\usepackage{amsmath}
\usepackage{algorithm}
\usepackage{algpseudocode}
\usepackage{amssymb}
\usepackage{amsthm}
\usepackage{subcaption}
\usepackage{graphicx} 
\usepackage[T1]{fontenc}
\usepackage[utf8]{inputenc}
\usepackage{microtype}
\usepackage{inconsolata}

\newtheorem{theorem}{Theorem}[section]

\newtheorem{definition}[theorem]{Definition}
\newtheorem{assumption}[theorem]{Assumption}
\newtheorem{theo}[theorem]{Theorem}

\newcommand{\ours}{\texttt{CCD}}
\newcommand{\oursdef}{\texttt{CCED}}

\title{LLM-Assisted Content Conditional Debiasing for Fair Text Embedding}
\begin{document}

\author{Wenlong Deng$^{1,2}$\thanks{Work done at google.}, Blair Chen$^{2}$, Beidi Zhao$^{1}$, Chiyu Zhang$^{1}$, \\
\textbf{Xiaoxiao Li}$^{1}$\thanks{Equal Corresponding Author.}, \textbf{Christos Thrampoulidis}$^{1\dag}$   \\
Department of Electrical and Computer Engineering$^1$,\\ The University of British Columbia, Vancouver, BC, Canada \\
\texttt{\{dwenlong,beidiz,chiyuzh,xiaoxiao.li,cthrampo\}@ece.ubc.ca} \\
Google Cloud$^2$, USA \\
\texttt{\{chenblair\}@google.com} 
}
\maketitle
\begin{abstract}
Mitigating biases in machine learning models has become an increasing concern in Natural Language Processing (NLP), particularly in developing fair text embeddings, which are crucial yet challenging for real-world applications like search engines. In response, this paper proposes a novel method for learning fair text embeddings. First, we define a novel content-conditional equal distance (\oursdef{}) fairness for text embeddings, ensuring content-conditional independence between sensitive attributes and text embeddings. Building on \oursdef{}, we introduce a content-conditional debiasing (\ours{}) loss to ensure that embeddings of texts with different sensitive attributes but identical content maintain the same distance from the embedding of their corresponding neutral text. Additionally, we tackle the issue of insufficient training data by using Large Language Models (LLMs) with instructions to fairly augment texts into different sensitive groups. Our extensive evaluations show that our approach effectively enhances fairness while maintaining the utility of embeddings. Furthermore, our augmented dataset, combined with the \oursdef{} metric, serves as an new benchmark for evaluating fairness.
\end{abstract}
\section{Introduction}

Embedding text into dense representations is a widely used technique in modern NLP, powering applications such as sentiment analysis~\cite{dang2020sentiment}, recommendation systems~\cite{zhang2016collaborative}, and search engines~\cite{palangi2016deep}. However, the extensive use of these embeddings introduces inherent biases that can affect various applications~\cite{packer2018text,baeza2018bias,zerveas2022mitigating,rabelo2022overview}. For instance, search engines~\cite{huang2020embedding} preprocess all text contents and search queries into embeddings to optimize storage and enable efficient similarity matching. These inherent biases in text embeddings can influence the calculation of embedding similarity, impacting the filtering of numerous documents to find pertinent ones. Moreover, text embeddings are directly employed in other applications such as zero-shot classification~\cite{yin2019benchmarking,radford2021learning} and clustering~\cite{john2023exploration}.
Unfortunately, various forms of biases, including gender, racial, and religious biases, have been identified in text embeddings generated by pre-trained language models (PLMs), as reported in several studies~\cite{bolukbasi2016man,nissim2020fair,liang2020towards,may2019measuring}. Consequently, attaining fairness in text embedding models is crucial.
 \begin{figure*}[t] 
\centering
\includegraphics[width=\linewidth]{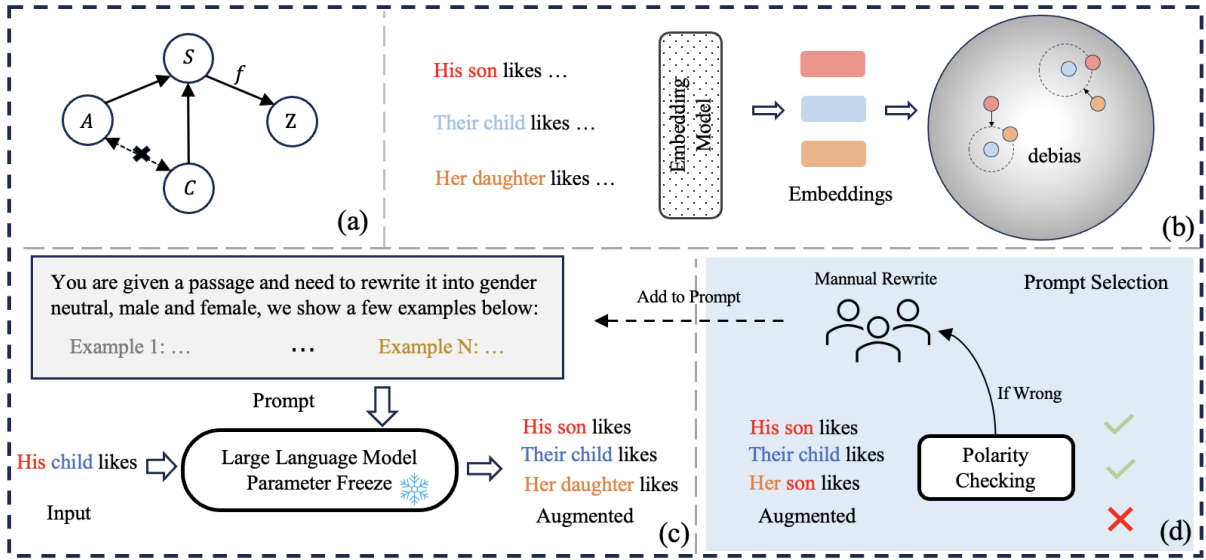} 
\caption{Pipleline of our method with \textbf{gender} as the sensitive attributes. 
(a) Graphical demonstration of the fairness issue. (b) The debiasing procedure achieves a content-conditioned equal distance to improve the fairness. (c) Overview of the data augmentation strategy, including the prompt template used to replace sensitive words with their equivalents from all sensitive groups.
(d) Prompt search module: Augmented texts are sent to the demographic polarity checking block. Incorrectly augmented samples are then manually labeled and added to the prompts.}
\label{fig:fair}
\vspace{-1mm}
\end{figure*}

Recent debiasing techniques~\cite{liang2020towards,kaneko2021debiasing} for text embeddings use post-training to address biases, avoiding the inefficiency of retraining sentence encoders for each new bias. When removing bias, projection-based methods~\cite{liang2020towards,kaneko2021debiasing} reduce an embedding’s projection onto each bias subspace. The distance-based method~\cite{yang2023adept} constructs embeddings for sensitive groups and equalizes distances to text embeddings across these groups. Nevertheless, these methods persist in pursuing independence between sensitive attributes and text embeddings, which results in the complete removal of sensitive information. As a result, these approaches do not effectively find the \textit{sweet spot} between fairness and utility trade-off~\cite{zhao2022inherent,deng2023fairness,zliobaite2015relation}. 

Recent studies~\cite{mary2019fairness,deng2023fairness,pogodin2022efficient} suggest that using datasets labeled with sensitive information to achieve conditional independence --- specifically, conditioning on the content class to preserve semantic information within the text — provides a more effective approach to achieving fairness while preserving utility. Yet, the scarcity of text datasets with sensitive labels~\cite{gallegos2023bias} limits the practical application of these findings. To create such datasets, Counterfactual Data Augmentation (CDA)~\cite{zhao2018gender} collects sensitive-related words and employs a rule-based method to augment the data, but this approach encounters challenges due to the need for an extensive list of words. Finally, while Large Language Models (LLMs)~\cite{schick2021generating,shao2023synthetic} have offered new methods for data generation thanks to their rich contextual knowledge, yet they still struggle with inherent systematic biases~\cite{yu2023large}.

In this paper, we improve the text embeding fairness through defining fairness with theoretical analysis, a novel debiasing loss design, and an LLM-based data strategy for dataset generation. Our contributions include:
\begin{itemize}[leftmargin=*]
    \item Introducing \oursdef{} fairness for text embeddings, ensuring equal sensitive information and conditional independence between sensitive attributes and embeddings.
    \item Proposing \ours{} loss to achieve the desired \oursdef{} fairness by ensuring that texts with varied sensitive attributes but identical content have embeddings equidistant from their neutral counterparts.
    \item Employing LLMs to augment datasets fairly, representing diverse sensitive groups within the same content for effective training with \ours{}. Proposing polarity-guided prompting to ensure the LLM-generated data quality and minimize the potential biases from LLMs. 
    \item Establishing \oursdef{} fairness as a benchmark for evaluating fairness in text embeddings.
    \item Extensive evaluations on debiasing benchmarks and downstream tasks demonstrate \ours{}'s effectiveness in promoting fairness while preserving utility.
\end{itemize}

\section{Related Work}
\textbf{Debias Text Embedding:} Bias in text embeddings (also known as sentence embedding) is a significant issue that arises when these models reflect or amplify societal stereotypes and prejudices found in their training data. To resolve the issue, ~\cite{liang2020towards} contextualizes predefined sets of bias attribute words to sentences and applies a hard-debias algorithm~\cite{bolukbasi2016man}. Contextualized debiasing methods~\cite{kaneko2021debiasing,yang2023adept} apply token-level debiasing for all tokens in a sentence and can be applied at token- or sentence-levels~\cite{kaneko2021debiasing} to debias pretrained contextualized embeddings. However, all the above methods aim to strictly achieve independence between text embedding and sensitive attributes, which may not balance fairness and utility well. While \citet{liu2021contrastive, shen-etal-2022-representational} employ contrastive learning losses to mitigate biases in language representations for text classification, their approach relies on supervised data, which is often scarce and expensive to obtain, and primarily focuses on fairness in the subsequent task. Additionally, although \cite{leteno2023fair, shen-etal-2022-representational} observe that representational fairness and group fairness in subsequent tasks are either not correlated or only partially correlated, it is important to note that fairness in subsequent tasks and fairness in text embeddings are distinct areas, with the latter being crucial for various applications. A detailed discussion of these differences can be found in Appendix~\ref{sec:diff_s_i}. In this paper, we utilize LLMs to augment training data for learning fair text embeddings with proposed \ours{} loss. 

\noindent\textbf{LLMs for Dataset Generation:} Leveraging the success of LLMs, researchers have begun using them to generate various forms of training data, such as tabular data~\cite{borisov2022language}, relation triplets~\cite{chia2022relationprompt}, sentence pairs~\cite{schick2021generating, zhang-etal-2024-distilling}, and instruction data~\cite{shao2023synthetic, wu-etal-2024-lamini}. As we focus on obtaining data with sensitive attribute information, data generation for text classification would be the most similar one among those applications. Recent efforts in generating data for text classification ~\cite{meng2022generating,ye2022zerogen,wang2019symmetric} primarily employ simple class-conditional prompts while focusing on mitigating issues of low quality after generation. However, these efforts encounter the challenge of inherent systematic biases present in LLMs~\cite{yu2023large}. While~\citet{yu2023large} considers generated data bias, it focuses only on the diversity of topics and overlooks the inherent bias within words in a text (e.g. `child' occurs more frequently with`mother'). In this paper, we instructs the LLM to only locate the gendered words and replace them with counterparts from other groups and propose polarity-guided prompt searching to minimize biases from LLMs and ensure the quality of augmented data.

\section{Method}
\subsection{Problem Setting}\label{sec:prob}

This section outlines the problem of fairness in text embeddings. We define several key variables: \( S \in \mathcal{D} \) represents the input text from the data distribution, \( C \) denotes the content of the text,\footnote{For instance, the texts `he is a teacher' and `she is a teacher' both convey the same content \( C \) = `is a teacher'.} and \( A = [a_1, \ldots, a_{|A|}] \) represents the sensitive attributes (e.g. gender and age). The symbol \( n \) indicates neutral, meaning no sensitive information is present. A text with content \( C \) is considered neutral \( S_C^n \) if it contain no sensitive information, whereas text \( S_C^{a_i} \) is associated with the sensitive attribute \( a_i \) if its sensitive polarity~\cite{wang2023toward} is \( a_i \), see Eq.~\eqref{eq:p_check}. The text embedding model \( f \) processes a text into a \( d \)-dimensional embedding \( Z \in \mathbb{R}^d \). The embedding of a neutral text encodes the content information $C'$ (a well trained model $C' \approx C$), while the embedding of a sensitive text additionally encodes sensitive information. Words in the text related to the attribute \( a_i \) are denoted as \( X^{a_i} \), and neutral words are denoted as \( X^n \). For clarity, we provide detailed notations in Table~\ref{tab:Notation} in Appendix.
\\
\textbf{Fairness Issue:} Fig.~\ref{fig:fair} (a) shows there exists an association between attributes $A$ and content variable $C$. If model $f$ superficially treats $A$ as a proxy for $C$\footnote{For instance, raising children is frequently associated with women in the training corpus, resulting in the proxy effect.}, it results in encoded $C'$ being represented by $A$ thus embedding $Z$ will mainly contain sensitive information, which leads to issues of fairness.
\\
\textbf{Fairness Goal:} Mitigating fairness is not trivial, as we need to address not only bias mitigation but also the protection of the model's representation ability.  As shown in Fig.~\ref{fig:fair} (a), our method aims to (1) break the association between content $C$ and the sensitive attribute $A$, and (2) preserve useful sensitive information in the text embedding. For example, in the case of a text about a father raising a child, its embedding should retain information about the father.
\subsection{Content Conditional Debiasing}
To break the superficial association, we propose to achieve conditional independence between sensitive attributes and content $A \perp C' ~|~ C$. The conditional independence allows prediction $C'$ to depend on $A$ but only through the content variable $C$, prohibiting abusing $A$ as a \textit{proxy} for $C$ thus mitigating the fairness issue while preserving the utility. To protect utility, our objective is not to completely remove sensitive information but to ensure that text embeddings from different sensitive groups with identical content contain an equal amount of sensitive information.
\subsubsection{Fairness Definition} Firstly, we propose a novel content conditional equal distance fairness for fair text embedding:
\begin{definition}(Content Conditional Equal Distance  (CCED) Fairness.)\label{def:fairness}
Let $S_C^n$ be a neutral text with content $C$. Assume $S_C^A = [S_C^{a_1}, S_C^{a_2},..., S_C^{a_{|A|}}]$ being a set of texts from all sensitive groups with the same content $C$. Then, embedding model $f$ is content conditioned equal distance fair with respect to attributes $A$, for any $a_i,a_j \in A$:
    \begin{align}
        \|f(S_C^{a_i}) - f(S_C^{n})\| = \|f(S_C^{a_j}) &- f(S_C^{n})\|,
    \end{align}
where $\|\cdot\|$ is $L_2$ norm.
\end{definition}

As shown in Fig.~\ref{fig:fair} (b), CCED fairness requires that texts with the same context from different sensitive groups have equal distance to their corresponding neutral text on the embedding space. This text embedding fairness definition has two merits: 
\\
\underline{Equal sensitive information:} The equal distance to the neutral embedding ensures an equitable encoding of sensitive information across diverse groups, allowing fair usage of sensitive information and preserving the utility of embeddings.
\\
\underline{Content Conditional Independent:} Echoing the methodologies in previous research~\cite{hinton2002stochastic,yang2023adept}, the conditional independence $A \perp C' ~|~ C$ can be represented as the \oursdef{} on the embedding space:
\begin{assumption} (Equal Probability)\label{the:ass1} Within a content $C$, the likelihood $P(a_i|C)$ on all sensitive attributes $a_i \in A$ is uniform $P(a_1|C) = ... = P(a_A|C)$.
\end{assumption}
\begin{theo}\label{the:theory} When the equal probability assumption holds, achieving content conditioned equal distance fairness is equivalent to achieving conditional independence between sensitive attributes and content $A \perp C' ~|~ C$.
\end{theo}
\noindent Assumption~\ref{the:ass1} is true for a fair dataset that has balanced texts from all groups within content $C$ (can be obtained through our data augmentation strategy in Section~\ref{sec:ccd}). Theorem~\ref{the:theory} demonstrates the merit of CCED fairness (Definition~\ref{def:fairness}) in achieving embedding fairness. Detailed proof can be found in Appendix~\ref{sec:proof}.

\subsubsection{Content Conditional Debiasing Loss}
Based on the defined CCED fairness, we design a loss function $L_{bias}$ that aims to mitigate biases while preserving the representation ability of PLMs. For a sample pair $[S_C^{a_1},...,S_C^{a_{|A|}},S_C^{n}]$ :
\begin{align}\label{lossb}
    L_{bias} = \sum_{i \in [A]} \sum_{j \not = i} & |dist(f(S_C^{a_i}),f(S_C^{n})) \nonumber   \\ &- dist(f(S_C^{a_j}),f(S_C^{n})) |,
\end{align}
where $dist(A,B) = \exp\left(- \frac{\lVert A - B \rVert^2}{2\rho^2} \right)$ measures the distance on the embedding manifold~\cite{yang2023adept,hinton2002stochastic} (details in Appendix~\ref{sec:proof}), and $\rho$ is selected as the variance of the distance over the training dataset for normalization. To further preserve the valuable information encoded in the model and achieve efficient debiasing, we design $L_{rep}$ to enforce high similarity between the neutral texts' embeddings processed by the fine-tuned model $f$ and those processed by the original model $f^{org}$:
\begin{align}\label{lossr}
L_{rep} = \|f(S^{n}) - f^{org}(S^n)\|.
\end{align}
Ensuring that neutral embeddings remain unchanged offers two benefits: preserving the model's representational capability and maintaining neutral embeddings as a consistent reference point in the debiasing loss, ensuring stable equal distance to embeddings with various sensitive attributes. Thus, the overall training objective is:
\begin{align}\label{lossall}
    L_{all} = L_{bias} + \beta * L_{rep},
\end{align}
where $\beta$ is a hyper-parameter used to balance the two terms. An ablation study for setting $\beta$ is detailed in Table~\ref{tab:beta_ab}.

\subsection{LLM-Assisted Content Conditional Data Augmentation}\label{sec:ccd} 
We leverage the rich contextual knowledge of LLM with few-shot prompting to obtain a dataset that 
(1) fulfills the Assumption~\ref{the:ass1} to achieve our goal in Definition~\ref{def:fairness} as well as (2) avoids introducing inherent bias in LLM to augmented data. The data augmentation algorithm is shown in Alg. \ref{alg:aug}, followed by a detailed explanation below.
\begin{algorithm}[t]
\caption{Data Augmentation Algorithm}\label{alg:aug}
\textbf{Input:} Dataset $\mathcal{D}$, Sensitive word lists $V$, Pretrained LLM $h$, Task Description $T$, Example Prompts $P$. 
\begin{algorithmic}[1]
\For{$k$ in $1, \ldots, K$} \Comment{$K=10$ in this work}
\State \textbf{Block I}: Augment Texts into Different Sensitive Groups
\For{$S \in \mathcal{D}$}
\State $h(S, T, P)\rightarrow[S^{a_1}, ..., S^{a_{|A|}}, S^n],c$
\EndFor
\If{$k=K$}
\State \textbf{return} Augmented Dataset $\mathcal{D}'$
\EndIf
\State \textbf{Block II}: Polarity Guided Prompt Searching
\For{$[S^{a_1}, ..., S^{a_{|A|}}, S^n] \in D'$}
\State  Polarity Checking Eq.\ref{eq:p_check}
\EndFor
\State Manually Augment the wrong augmentation with highest $c$ and add to $P$.
\EndFor
\end{algorithmic}
\end{algorithm}

\noindent\textbf{Augment Text into Different Sensitive Groups:}
As shown in Fig.~\ref{fig:fair} (c), our task description $T$ instructs the LLM to only locate the gendered words and replace them with counterparts from other groups, leaving the other content unchanged thus avoiding fairness issues in text generation. Specifically, for sensitive words $X^A = [X^{a_i}, ..., X^{a_j}], a_i, a_j \in A$ in the text $S$, the LLM $h$ substitutes $X^A$ with words from different sensitive groups and neutral terms, thus obtaining augmented texts from all sensitive groups (as shown in Table~\ref{tab:generated}):
\begin{align}
h(S, T, P) = [S^{a_1}, ..., S^{a_{|A|}}, S^n],c
\end{align}
where  $c$ is the confidence score and $P$ is the example prompts (detailed in Table~\ref{tab:task_template} in Appendix). After augmentation, the dataset will have an equal amount of texts from each sensitive group with identical content, meeting our equal probability Assumption~\ref{the:ass1}. 
 \\
\noindent\textbf{Polarity-Guided Prompt Searching:} To ensure the quality of augmented texts and the effectiveness of few-shot prompt tuning on LLMs, finding appropriate prompts \(P\) is crucial. We propose identifying difficult samples from incorrectly augmented texts to use as prompts. First, these incorrectly augmented samples are detected through a sensitive polarity check as described by~\cite{wang2023toward} and illustrated in Fig.~\ref{fig:fair}(d). By counting the occurrences of words in predefined sensitive word lists \(V = [V^{a_i},...,V^{a_j}], a_i, a_j \in A\), the polarities of a series of sentences are determined as follows:
\begin{align}\label{eq:p_check}
    g(S) = \arg\max_{a_i\in A} occ(S, V^{a_i}),
\end{align}
where \(occ\) represents the number of times words from the list \(V^{a_i}\) appear in all augmented sentences \(S\). For a properly augmented sentence \(S^{a_i}\), its polarity should match the sensitive attribute \(a_i\). If \(g(S^{a_i}) \neq a_i\), the sentence is considered inaccurately augmented. Then we introduce our prompt searching strategy in Algorithm~\ref{alg:aug}. In each iteration, the algorithm identifies the incorrectly augmented sample with the highest confidence \(c\), manually augments it, and adds it to the example prompts \(P\). This rule-guided prompt search is repeated \(K\) times (with \(K=10\)) to prepare samples for the few-shot prompt tuning of de-biasing LLMs.
\\

\begin{table*}
\resizebox{\linewidth}{!}{%
 \begin{tabular}{c|c}
 \hline \hline
 Gender & Generated Text \\
  \hline
 \multirow{4}{*}{Male} & 
But because \textcolor{red}{Rumsfeld} wanted to prove a point about transforming strategy.\\
& After championing the continuation of \textcolor{red}{his} hardline policy, \textcolor{red}{his} current strategy of negotiation is risky. \\
&  \textcolor{red}{He} has been very vocal in voicing discontent with the rule of Kirchner and that of \textcolor{red}{his husband} and predecessor, Néstor Kirchner.
\\
\hline
 \multirow{4}{*}{Neutral} & 
But because the \textcolor{blue}{individual} wanted to prove a point about transforming strategy. \\
& After championing the continuation of \textcolor{blue}{their} hardline policy, \textcolor{blue}{the} current strategy of negotiation is risky. \\
& \textcolor{blue}{They} have been very vocal in voicing discontent with the rule of Kirchner and that of \textcolor{blue}{their spouse} and predecessor, Néstor Kirchner. \\
\hline
 \multirow{4}{*}{Female} & 
But because \textcolor{orange}{Rachel} wanted to prove a point about transforming strategy. \\
& After championing the continuation of \textcolor{orange}{her} hardline policy, \textcolor{orange}{her} current strategy of negotiation is risky. \\
& \textcolor{orange}{She} has been very vocal in voicing discontent with the rule of Kirchner and that of \textcolor{orange}{her wife} and predecessor, Néstor Kirchner.  \\
\hline \hline
\end{tabular} \\
}
\caption{We utilize LLM to augment text into three gender categories: Male, Female, and Neutral. Below are sample examples of the generated data, where words containing gender information are highlighted in colors: \textcolor{red}{red} for male, \textcolor{blue}{blue} for neutral, and \textcolor{orange}{orange} for female.}
\label{tab:generated}
\vspace{-3mm}
\end{table*}

\section{Experiments}
In this paper, we take \textit{gender} bias as an example due to its broad impact on society.

\noindent\textbf{Datasets:} We utilize the News-commentary-v15 corpus~\cite{tiedemann2012parallel} as source samples to generate our training data with LLMs. For gender bias evaluation, we follow~\cite{yang2023adept} to use SEAT~\cite{may2019measuring}, CrowS-Pairs~\cite{nangia2020crows} and StereoSet-Intrasentence data~\cite{nadeem2020stereoset}. We additionally assess fairness on longer texts via the Bias-IR dataset~\cite{krieg2023grep}. To evaluate whether the biased models’ representation ability is maintained, we follow~\cite{kaneko2021debiasing,yang2023adept} to select four small-scale subsequent tasks from the GLEU benchmark: Stanford Sentiment Treebank (SST-2~\cite{socher-etal-2013-recursive}), Microsoft Research Paraphrase Corpus (MRPC~\cite{dolan-brockett-2005-automatically}), Recognizing Textual Entailment (RTE~\cite{bentivogli2009fifth}) and Winograd Schema Challenge (WNLI~\cite{levesque2012winograd}). More dataset information see Appendix~\ref{sec:datadetail}.

\noindent\textbf{Backbone and Baseline Methods:}
For the selection of PLMs, we choose BERT-large-uncased~\cite{devlin2018bert} and RoBERTa-base~\cite{liu2019roberta}.
To assess debiasing performance, we compare our algorithm with finetuning-based methods DPCE~\cite{kaneko2021debiasing} and ADEPT-F~\cite{yang2023adept}.
To assess the effectiveness of our data augmentation strategy, we compare our approach with CDA~\cite{zhao2018gender}. 

\noindent\textbf{LLM-Assisted Data Augmentation:} We leverage ChatGPT (i.e., \texttt{gpt-3.5-tubo}) and Gemini~\cite{team2023gemini} to generate our training data. We obtained a dataset with texts of content $C$ from all groups $A$ and neutral. Using Gemini and ChatGPT for data augmentation resulted in datasets with 43,221 and 42,930 sample pairs, respectively. Examples of data augmented through our method are presented in Table~\ref{tab:generated}, and the quality of the augmented dataset is assessed in Section~\ref{sec:fair_acc}.

\noindent\textbf{Hyperparameters:}
We use Adam to optimize the objective function. During the debiasing training, our learning rate is 5e-5, batch size is 32, and $\beta$ is 1. Our method requires training for only a single epoch and selecting the checkpoint with the lowest validation loss (validate every 500 steps). The results for DPCE and ADEPT-F are obtained using the originally reported hyperparameters from the studies by~\cite{kaneko2021debiasing,yang2023adept}. Consistent with these studies, we set the random seed to 42 to ensure a fair comparison. All experiments are conducted on an NVIDIA A100 GPU.

\subsection{Augmentation Quality Checking}\label{sec:fair_acc}
To demonstrate the quality of our augmented data on gender, we quantitatively assess the fairness of our augmented dataset using the union gender polarity accuracy metric, formulated as follows:
\begin{align}
g_i^u = \big(g(S_i^n) = n &\cap g(S_i^m) = a_m \cap g(S_i^f) = a_f\big) \nonumber \\
Acc &= \frac{\sum_{i=1}^{N} g_i^u}{N},
\end{align}
where $[S_i^n, S_i^m, S_i^f]$ are the augmented texts for the $i$-th sample, $N$ denotes the size of the augmented dataset, and $g(\cdot)$ is the polarity checking function as defined in Eq.~\eqref{eq:p_check}. The union gender polarity accuracy metric measures the proportion of text triples (neutral, male, female) that are accurately augmented in alignment with their respective gender polarities. The results show both Gemini and GPT models achieve high accuracy, with Gemini and GPT reaching 83.4\% and 82.2\% respectively . This suggests that our data augmentation process has effectively produced a fair dataset. Incorporating polarity checking as a post-processing step further ensures the fairness of our augmented data.

\begin{table*}[t!]
\centering
\resizebox{\linewidth}{!}{
\begin{tabular}{lcccccccccccccc}
\hline \hline \multirow{2}{*}{\begin{tabular}{l} 
Datasets \\
Method
\end{tabular}} & \multicolumn{7}{c}{ SEAT $(0.00)$ the best } & \multicolumn{3}{c}{StereoSet:gender} & \multicolumn{3}{c}{StereoSet:all}  & \multicolumn{1}{c}{ CrowS-Pairs}\\ \cmidrule(l){2-8} \cmidrule(l){9-11} \cmidrule(l){12-14} \cmidrule(l){15-15}
 & $6$ & $6$-b & $7$ & $7$-b & $8$ & $8$-b & AVG (abs)$\downarrow$  & LMS$\uparrow$ & SS $(50.00)$ & ICAT$\uparrow$ & LMS$\uparrow$ & SS$(50.00)$ & ICAT$\uparrow$ & SS$(50.00)$ \\
\hline
BERT & $0.37$ & $0.20$ & $0.42$ & $0.22$ & $-0.26$ & $0.71$ & $0.36$ & $86.34$ & $59.66$ & $69.66$ & $84.16$ & $58.24$ & $70.29$ & $55.73$\\
DPCE & $-0.21$ &  $0.27$  &$0.44$ & $0.07$ & $0.25$ &$0.21$ &  $0.24$ &$81.19$ & $56.72$ & $65.41$ & $64.06$ & $\underline{52.96}$ & $ 60.26$ & $52.29$ \\
ADEPT-F & $0.83$ & $-0.14$ & $0.63$ & $1.24$ & $0.43$ & $1.28$ & $0.76$ & $86.45$ & $61.70$ & $66.21$ & $85.09$ & $57.52$ & $72.26$ & $51.91$ \\
\hline
DPCE-Gemini & $-0.63$ &  $0.41$  &$0.00$ & $-0.01$ & $0.19$ &$0.17$ &  $\textbf{0.23}$ &$82.63$ & $60.68$ & $64.98$ & $64.08$ & $54.91$ & $ 57.78$ & $51.53$  \\
ADEPT-F-Gemini & $0.71$ &  $-0.23$  &$0.21$ & $0.92$ & $0.35$ &$0.99$ &  $0.57$ &$86.80$ & $61.72$ & $66.44$ & $85.47$ & $58.50$ & $ 71.71$ & $51.91$ \\
\hline
\ours{}-CDA  & $0.16$ & $0.03$  & $0.43$ & $0.38$ & $0.47$  & $0.22$ & $0.29$ & $80.34$ & $\mathbf{53.53}$ & $74.69$ & $79.10$ & $53.46$ & $73.62$ & $46.95$\\
\ours{}-GPT  & $0.35$ & $-0.11$  & $-0.17$ & $-0.15$ & $0.57$  & $0.06$ & $\textbf{0.23}$ & $81.47$ & $\underline{53.60}$ & $\mathbf{75.60}$ & $80.22$ & $\textbf{52.83}$ & $\mathbf{75.97}$ & $\underline{47.71}$\\
\ours{}-Gemini & $0.47$ & $-0.00$  & $-0.02$ & $-0.72$ & $-0.30$  & $0.07$ & $\underline{0.26}$ & $82.91$ & $54.93$ & $\underline{74.72}$ & $82.97$ & $55.00$ & $\underline{74.67}$ & $\textbf{48.85}$\\
\hline\hline
\end{tabular}}
\caption{Comparison of debiasing performance on BERT. We test the debiased models on SEAT, CrowS-Pairs, and filtered StereoSet-Intrasentence, with the best and second best results in \textbf{bold} and \underline{underline} respectively.}
\label{tab:fairness}
\end{table*}

\subsection{Results and Analysis}


\begin{table*}[t!]
\centering
\resizebox{\linewidth}{!}{
\begin{tabular}{lcccccccccccccc}
\hline \hline \multirow{2}{*}{\begin{tabular}{l} 
Datasets \\
Method
\end{tabular}} & \multicolumn{7}{c}{ SEAT $(0.00)$ the best } & \multicolumn{3}{c}{StereoSet:gender} & \multicolumn{3}{c}{StereoSet:all}  & \multicolumn{1}{c}{ CrowS-Pairs}\\
\cmidrule(l){2-8} \cmidrule(l){9-11} \cmidrule(l){12-14} \cmidrule(l){15-15} 
& $6$ & $6$-b & $7$ & $7$-b & $8$ & $8$-b & AVG (abs)$\downarrow$  & LMS$\uparrow$ & SS $(50.00)$ & ICAT$\uparrow$ & LMS$\uparrow$ & SS$(50.00)$ & ICAT$\uparrow$ & SS$(50.00)$ \\
\hline
RoBERTa & $0.92$ & $0.21$&$0.98$&$1.46$&$0.81$&$1.26$&$0.94$ & $89.79$ & $66.17$ & $60.74$ & $88.91$ &$62.22$&$67.17$&$60.15$ \\
DPCE & $0.40$ &$0.11$&$0.73$&$0.98$ &$0.03$ & $0.75$&$0.50$& $82.93$ & $61.80$ & $64.11$ & $61.30$ & $55.14$&$54.99$&$54.79$ \\
ADEPT-F & $1.23$ & $-0.14$ & $0.99$ & $1.09$ & $0.93$ & $1.11$ & $0.92$ & $89.81$ & $63.10$ & $66.27$ & $90.03$ & $61.31$ & $69.68$ & $55.56$ \\
\hline
\ours{}-CDA  & $0.29$ & $-0.07$  & $0.87$ & $0.94$ & $0.58$  & $0.85$ & $0.60$ & $88.52$ & $60.29$ & $\underline{70.29}$ & $88.88$ & $59.12$ & $72.66$ & $\underline{50.57}$\\
\ours{}-GPT  & $0.40$ & $0.08$  & $0.41$ & $0.85$ & $0.57$  & $0.63$ & $\underline{0.49}$ & $87.21$ & $\underline{59.51}$ & $\textbf{70.63}$ & $88.33$ & $\underline{57.61}$ & $\textbf{74.89}$ & $48.66$\\
\ours{}-Gemini & $0.27$&$-0.18$&$-0.13$ & $0.82$&$0.08$&$0.81$&$\textbf{0.38}$&$81.35$&$\textbf{58.15}$&$68.10$&$84.68$&$\textbf{56.65}$&$\underline{73.41}$&$\textbf{49.54}$ \\
\hline\hline
\end{tabular}
}
\caption{Comparison of debiasing performance on RoBERTa. We test the debiased models on SEAT, CrowS-Pairs, and filtered StereoSet-Intrasentence, with the best and second best results in \textbf{bold} and \underline{underline} respectively.}
\label{tab:fairness_roberta}
\end{table*}

\begin{table*}[t!]
\centering
\resizebox{\linewidth}{!}{
\begin{tabular}{lccccccccccccc}
\hline \hline \multirow{2}{*}{\begin{tabular}{l} 
Datasets \\
Method
\end{tabular}}  &  \multicolumn{5}{c}{ GLUE $\uparrow$} & \multicolumn{8}{c}{ Bias-IR (Male Ratio, 0.50 the best)}\\ \cmidrule(l){2-6} \cmidrule(l){7-14} 
& SST-2$\uparrow$ & MRPC$\uparrow$ & RTE$\uparrow$ & WNLI$\uparrow$ & AVG$\uparrow$  & Appearance & Child & Cognitive & Domestic & Career & Physical & Relationship & AVG-DEV$\downarrow$ \\
\hline
BERT &  $92.9$& $84.6$ & $\underline{72.5}$ & $38.0$ & $72.0$ &
$0.71$ & $0.50$ & $0.75$ & $0.46$ & $0.75$ & $0.68$ & $0.61$  & $0.16$
\\
DPCE & $92.8$ & $69.6$ & $53.4$ & $49.3$ & $66.3$ 
& $0.86$ & $0.79$ & $1.00$ & $0.47$& $0.70$ & $0.84$ & $0.61$& $0.24$
\\
ADEPT-F & $93.2$ & $\underline{85.5}$ & $69.9$ & $56.3$ & $76.2$ 
& $0.50$ & $0.50$  & $0.75$ &$0.53$ & $0.80$ & $0.68$  & $0.65$  & $0.13$
\\
\hline
DPCE-Gemini & $93.2$ & $81.4$ & $60.6$ & $46.5$ & $70.4$ 
& $0.29$ & $0.36$ & $0.17$ & $0.20$& $0.10$ & $0.32$ & $0.35$  & $0.24$
\\
ADEPT-F-Gemini & $92.7$ & $81.4$ & $71.5$ & $56.3$ & $75.5$ 
& $0.71$ & $0.43$  & $0.83$ &$0.53$ & $0.65$ & $0.74$  & $0.65$ & $0.17$
\\
\hline
\ours{}-CDA & $92.8$ & $\textbf{86.3}$ & $65.3$ & $50.7$ & $73.8$ 
& $0.79$ & $0.79$ & $0.83$ & $0.80$ & $0.70$ & $0.79$ &  $0.83$ & $0.29$ \\
\ours{}-GPT &  $\textbf{93.6}$ & $85.1$  & $70.4$ & $\underline{56.3}$ & $\underline{76.4}$ & $0.78$ & $0.78$ & $0.50$ & $0.73$ & $0.50$ & $0.63$ &  $0.52$ &  $\underline{0.13}$\\
\ours{}-Gemini  & $\underline{93.5}$ & $83.6$ & $\textbf{72.9}$ & $\textbf{56.3}$  & $\textbf{76.6}$
& $0.57$ & $0.64$ & $0.58$ & $0.60$ &  $0.70$ &  $0.42$  & $0.65$  & $\textbf{0.11}$ \\


\hline\hline
\end{tabular}
}
\caption{Evaluation results on the GLUE dataset and the Bias-IR dataset with BERT, we calculate the average deviation to 0.5 for Bias-IR as AVG-DEV. The \textbf{bold} and \underline{underline} represent the best and second-best respectively.}
\label{tab:perf}
\end{table*}

\begin{table*}[t!]
\centering
\resizebox{\linewidth}{!}{
\begin{tabular}{lccccccccccccc}
\hline \hline \multirow{2}{*}{\begin{tabular}{l} 
Datasets \\
Method
\end{tabular}}  &  \multicolumn{5}{c}{ GLUE $\uparrow$} & \multicolumn{8}{c}{ Bias-IR (Male Ratio, 0.50 the best)}\\ \cmidrule(l){2-6} \cmidrule(l){7-14} 
& SST-2$\uparrow$ & MRPC$\uparrow$ & RTE$\uparrow$ & WNLI$\uparrow$ & AVG$\uparrow$  & Appearance & Child & Cognitive & Domestic & Career & Physical & Relationship & AVG-DEV$\downarrow$ \\
\hline
RoBERTa &  $93.8$& $88.2$ & $70.8$ & $56.3$ & $76.9$ &
$0.28$ & $0.28$ & $0.66$ & $0.40$ & $0.60$ & $0.42$ & $0.70$  & $0.16$
\\
DPCE & $78.1$& $81.6$ & $53.8$ & $56.3$ & $67.5$
& $0.43$ & $0.93$ & $0.42$ & $0.60$& $0.50$ & $0.58$ & $0.43$& $0.12$
\\
ADEPT-F & $93.9$ & $\textbf{89.2}$ & $66.8$ & $56.3$ & $76.6$ 
& $0.57$ & $0.50$  & $0.83$ &$0.60$ & $0.85$ & $0.68$  & $0.74$  & $0.18$
\\
\hline
\ours{}-CDA & $\underline{94.3}$ & $\underline{88.2}$ & $68.2$ & $56.3$ & $76.7$ 
& $0.29$ & $0.50$ & $0.58$ & $0.13$ & $0.35$ & $0.21$ &  $0.56$ & $0.16$ \\
\ours{}-GPT &  $93.1$ & $86.5$  & $\underline{71.5}$ & $56.3$ & $\underline{76.9}$ & $0.43$ & $0.36$ & $0.58$ & $0.33$ & $0.55$ & $0.53$ &  $0.61$  &  $\textbf{0.09}$\\
\ours{}-Gemini  & $\textbf{94.6}$ & $86.5$ & $\textbf{72.9}$ & $56.3$  & $\textbf{77.6}$
& $0.43$ & $0.50$ & $0.67$ & $0.53$ &  $0.65$ &  $0.58$  & $0.69$ & $\underline{0.10}$ \\
\hline\hline
\end{tabular}
}
\caption{Evaluation results on the GLUE dataset and the Bias-IR dataset with RoBERTa, we calculate the average deviation to 0.5 for Bias-IR as AVG-DEV. The \textbf{bold} and \underline{underline} represent the best and second-best respectively.}
\label{tab:perf_v2}
\end{table*}
\begin{table}[h]
\centering
\begin{subtable}{.5\linewidth}
\centering
\resizebox{\linewidth}{!}{
\begin{tabular}{lc}
\hline\hline
\multicolumn{1}{c}{Method} & CCED $\downarrow$ \\ \cline{1-2} 
\hline\hline 
BERT & 0.339 \\
DPCE & 0.212 \\
ADEPT-F & 0.324 \\ \hline
\ours{}-CDA & 0.081 \\
\ours{}-GPT & \textbf{0.056} \\
\ours{}-Gemini & \underline{0.077} \\
\hline\hline 
\end{tabular}
}
\caption{CCED on BERT.}
\end{subtable}%
\begin{subtable}{.5\linewidth}
\centering

\resizebox{\linewidth}{!}{
\begin{tabular}{lc}
\hline\hline
\multicolumn{1}{c}{Method} & CCED $\downarrow$ \\ \cline{1-2}
\hline\hline
RoBERTa & 0.438 \\
DPCE & 0.177 \\
ADEPT-F & 0.159 \\ \hline
\ours{}-CDA  & 0.166 \\
\ours{}-GPT & \underline{0.143} \\
\ours{}-Gemini & \textbf{0.052} \\
\hline\hline 
\end{tabular}
}
\caption{CCED on RoBERTa.}
\end{subtable} 
\caption{Debiasing performance in terms of CCED.}
\vspace{-5mm}
\label{tab:CCED}
\end{table}
We evaluate four models on all benchmarks, namely the original model (pre-trained with no explicit debiasing), the DPCE model, the ADEPT-F model, and our \ours{}. 

\noindent\textbf{Reducing Gender Biases:}
In Table~\ref{tab:fairness} and Table~\ref{tab:fairness_roberta}, our experiments demonstrate that \ours{} with GPT and Gemini data strategies excels in debiasing, consistently outperforming baselines in the StereoSet and CrowS-Pairs datasets for both BERT and RoBERTa backbones. On SEAT, both \ours{} and DPCE achieve good performance, with \ours{}-Gemini achieving the best overall performance on SEAT across both backbones. Notably, our method attains a high ICAT score in the StereoSet dataset, indicating an excellent balance between performance and fairness. However, while DPCE maintains great fairness, it adversely affects its representation capability, as evidenced by a significantly lower LMS score in the StereoSet dataset.

\noindent\textbf{Preserving Representation Ability:} In Table~\ref{tab:perf} and Table~\ref{tab:perf_v2}, the GLUE results demonstrate that \ours{}-Gemini achieves the highest average performance with both BERT and RoBERTa backbones, suggesting that our \ours{} even enhances the model's representation capabilities. Conversely, DPCE, which strictly separate gender attributes from neutral text embeddings, harms the model's utility.

\noindent\textbf{Bias in Information Retrieval:}
Since search engine performance is a crucial subsequent task of text embedding usage, we evaluate the bias in information retrieval using the Bias-IR dataset. For the BERT model, Table~\ref{tab:perf} shows that \ours{}-Gemini achieves the best fairness, with \ours{}-GPT ranking second. For the RoBERTa model, Table~\ref{tab:perf_v2} demonstrates \ours{}-GPT achieves the best fairness, with \ours{}-Gemini ranking second. Overall, \ours{} with GPT and Gemini data strategies outperforms baselines in fairness across various fields, as well as in average fairness.

\begin{figure*}[t] 
\centering
\includegraphics[width=\linewidth]{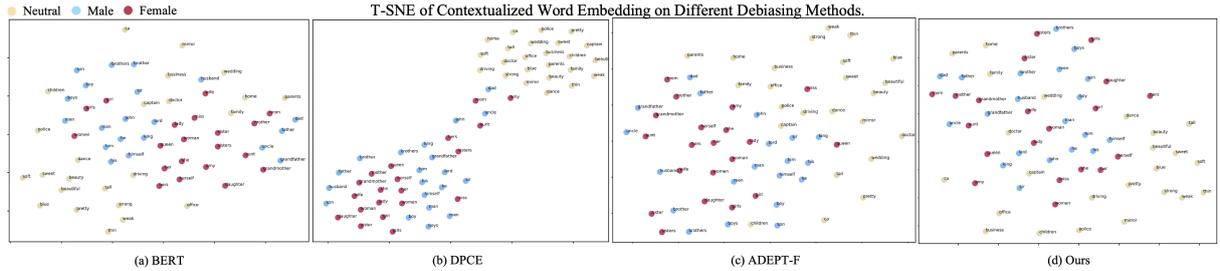} 
\caption{T-SNE plots of embeddings that are processed by different methods. Our approach maintains embedding positions similar to BERT while mixing male and female embeddings thus achieving fairness.}
\label{fig:tsne}
\vspace{-2mm}
\end{figure*}

\noindent\textbf{\oursdef{} as Fairness Metric:} We use our \oursdef{} fairness from Definition~\ref{def:fairness} to evaluate fairness. Specifically, we calculate the \oursdef{} gap for all methods on our Gemini-augmented dataset using the equation $\frac{1}{N}\sum^{N}_{i}|\|f(S_{i}^{a_i}) - f(S_{i}^{n})\| - \|f(S_{i}^{a_j}) - f(S_{i}^{n})\||$. Table~\ref{tab:CCED} demonstrates that \ours{} achieves the best fairness on the \oursdef{} fairness metric and  DPCE being the fairest baseline. The \oursdef{} results align well with the results on other benchmarks in Table~\ref{tab:fairness} and Table~\ref{tab:fairness_roberta}, indicating that \oursdef{} serves as an new benchmark for text embedding fairness.

\noindent\textbf{Comparision of Data Strategy:}
To demonstrate the effectiveness of our proposed data strategy, we conduct comparisons with CDA as shown in Table~\ref{tab:fairness} to Table~\ref{tab:perf_v2}. Integrating our debiasing loss with all data strategies results in improved fairness. However, CDA consistently performs worse than GPT and Gemini on fairness due to its limited sensitive word list. This highlights the superiority of our LLM-based augmentation method in leveraging the rich contextual knowledge of LLMs. For the use of different LLMs, both ChatGPT and Gemini achieve strong performance.
 
\noindent\textbf{Baseline with augmented data:}
In this section, we study of baseline methods with our Gemini augmented data and denote as DPCE-Gemini and ADEPT-F-Gemini . Table~\ref{tab:fairness} shows that our augmented dataset marginally improves fairness in certain metrics, though the overall performance remains similar to that of the original dataset. We arrive at the same conclusion: our \ours{} surpasses these baseline approaches. Regarding representation capability and BiasIR performance, the results are reported in Table~\ref{tab:perf}. We observed that DPCE experienced an improvement in GLUE average performance, while ADEPT-F showed a slight decline. Despite these variations, both DPCE-Gemini and ADEPT-F-Gemini still exhibit a significant performance gap compared to \ours{} methods, as detailed in Table~\ref{tab:perf}. To summarize, even with our augmented dataset, our \ours{} still outperforms baseline methods.

\noindent\textbf{Influence of $\beta$:}
We perform the ablation study of $\beta$ on \ours{}-Gemini using the StereoSet dataset on BERT, known for its comprehensive evaluation metrics that assess performance (LMS), fairness (SS), and the trade-off between them (ICAT). We highlight that increasing \(\beta\) amplifies the impact of the $L_{rep}$, as detailed in Eq.~\ref{lossall}, ensuring that neutral embeddings remain unchanged. This provides two key benefits: preserving the model's representational capability and maintaining neutral embeddings as a consistent reference point in the debiasing loss. We vary \(\beta\) from 0 to 1.5, with the results presented in the Table~\ref{tab:beta_ab}.
\begin{table}[t]
    \centering
    \begin{tabular}{ccccc}
        \hline
        \hline
        Method & $\beta$ & LMS  & SS   & ICAT \\
        \hline
        \multirow{4}{*}{CCD-Gemini} & 0.0     & 64.37 & \textbf{51.03} & 63.02 \\
                   & 0.5   & 73.67 & 53.69 & 68.22 \\
                   & 1.0   & 82.91 & 54.93 & \textbf{74.72} \\
                   & 1.5   & \textbf{84.28} & 57.64 & 71.39 \\
        \hline
         \hline
    \end{tabular}
    \caption{Influence of $\beta$ on StereoSet dataset with BERT.}
    \label{tab:beta_ab}
    \vspace{-3mm}
\end{table}
As $\beta$ increased, we observed an increase in the LMS score from 64.37 to 84.28, indicating improved model utility. However, the fairness score decreased from 57.64 to 51.03, suggesting a shift towards prioritizing utility over fairness. Setting $\beta=1$ resulted in the optimal ICAT score, balancing fairness and utility.

\noindent\textbf{Embedding Visualization:}
(1) Fairness Improvement: Fig.~\ref{fig:tsne}.a shows the T-SNE of the original BERT model, where male (blue dots) and female (red dots) embeddings form distinct clusters, indicating fairness issues~\cite{peltonen2023fair}. In contrast, baseline methods and our \ours{} mix male and female embeddings, thus improving fairness.
(2) Utility Preservation: DPCE (Fig.~\ref{fig:tsne}.b) separates gendered (blue and red) and neutral (yellow) embeddings, completely removing sensitive information. This disrupts the original embedding geometry and significantly reduces performance (Tables~\ref{tab:fairness} and \ref{tab:perf}). ADEPT (Fig.~\ref{fig:tsne}.c) also causes a performance drop and worsens fairness, as shown in Tables~\ref{tab:fairness} and \ref{tab:perf}. Notably, our approach (Fig.~\ref{fig:tsne}.d) maintains an embedding geometry similar to BERT while mixing male and female embeddings, achieving fairness without compromising utility.

\section{Conclusion}
In conclusion, we introduce \oursdef{} fairness for text embeddings, ensuring conditional independence and equal sensitive information between attributes and embeddings. We propose the \ours{} loss to achieve this fairness by ensuring that texts with varied sensitive attributes but identical content have equidistant embeddings from their neutral counterparts. By employing LLMs to fairly augment datasets, we achieve effective training with \ours{}. We establish \oursdef{} fairness as a benchmark for evaluating text embeddings fairness. Extensive evaluations on debiasing benchmarks and downstream tasks demonstrate \ours{}'s effectiveness in promoting fairness while preserving utility.

\section{Limitaions}
In this study, we utilize gender bias to demonstrate the efficacy of our method. As our approach constitutes a general pipeline, we plan to extend our methodology to address other types of biases (e.g., race, age) in the future. Moreover, we discuss the application of our method in a binary gender setting, which generally does not reflect the real world where gender (and other biases) may not be strictly binary. Fortunately, our method is readily extensible to any number of dimensions. We consider this extension as part of our future work.
\section{Ethical Consideration}
Our work pioneers in mitigating biases in text embeddings, crucial for fairness and inclusivity in NLP applications. We introduce a method that ensures fair representation by achieving conditional independence between sensitive attributes and text embeddings, aiming to reduce societal biases. Employing LLMs for data augmentation represents ethical advancement in tackling inherent biases, moving towards equitable technology and inspiring future bias-aware research. Our contribution significantly advances AI fairness by validating a method that minimizes bias in text embeddings, promoting inclusivity in machine learning.
\bibliography{acl_latex}

\begin{thebibliography}{53}
\expandafter\ifx\csname natexlab\endcsname\relax\def\natexlab#1{#1}\fi

\bibitem[{Baeza-Yates(2018)}]{baeza2018bias}
Ricardo Baeza-Yates. 2018.
\newblock Bias on the web.
\newblock \emph{Communications of the ACM}, 61(6):54--61.

\bibitem[{Bentivogli et~al.(2009)Bentivogli, Clark, Dagan, and Giampiccolo}]{bentivogli2009fifth}
Luisa Bentivogli, Peter Clark, Ido Dagan, and Danilo Giampiccolo. 2009.
\newblock The fifth pascal recognizing textual entailment challenge.
\newblock \emph{TAC}, 7(8):1.

\bibitem[{Bolukbasi et~al.(2016)Bolukbasi, Chang, Zou, Saligrama, and Kalai}]{bolukbasi2016man}
Tolga Bolukbasi, Kai-Wei Chang, James~Y Zou, Venkatesh Saligrama, and Adam~T Kalai. 2016.
\newblock Man is to computer programmer as woman is to homemaker? debiasing word embeddings.
\newblock \emph{Advances in neural information processing systems}, 29.

\bibitem[{Borisov et~al.(2022)Borisov, Se{\ss}ler, Leemann, Pawelczyk, and Kasneci}]{borisov2022language}
Vadim Borisov, Kathrin Se{\ss}ler, Tobias Leemann, Martin Pawelczyk, and Gjergji Kasneci. 2022.
\newblock Language models are realistic tabular data generators.
\newblock \emph{arXiv preprint arXiv:2210.06280}.

\bibitem[{Chia et~al.(2022)Chia, Bing, Poria, and Si}]{chia2022relationprompt}
Yew~Ken Chia, Lidong Bing, Soujanya Poria, and Luo Si. 2022.
\newblock Relationprompt: Leveraging prompts to generate synthetic data for zero-shot relation triplet extraction.
\newblock \emph{arXiv preprint arXiv:2203.09101}.

\bibitem[{Dang et~al.(2020)Dang, Moreno-Garc{\'\i}a, and De~la Prieta}]{dang2020sentiment}
Nhan~Cach Dang, Mar{\'\i}a~N Moreno-Garc{\'\i}a, and Fernando De~la Prieta. 2020.
\newblock Sentiment analysis based on deep learning: A comparative study.
\newblock \emph{Electronics}, 9(3):483.

\bibitem[{Deng et~al.(2023)Deng, Zhong, Dou, and Li}]{deng2023fairness}
Wenlong Deng, Yuan Zhong, Qi~Dou, and Xiaoxiao Li. 2023.
\newblock On fairness of medical image classification with multiple sensitive attributes via learning orthogonal representations.
\newblock In \emph{International Conference on Information Processing in Medical Imaging}, pages 158--169. Springer.

\bibitem[{Devlin et~al.(2018)Devlin, Chang, Lee, and Toutanova}]{devlin2018bert}
Jacob Devlin, Ming-Wei Chang, Kenton Lee, and Kristina Toutanova. 2018.
\newblock Bert: Pre-training of deep bidirectional transformers for language understanding.
\newblock \emph{arXiv preprint arXiv:1810.04805}.

\bibitem[{Dolan and Brockett(2005)}]{dolan-brockett-2005-automatically}
William~B. Dolan and Chris Brockett. 2005.
\newblock \href {https://aclanthology.org/I05-5002} {Automatically constructing a corpus of sentential paraphrases}.
\newblock In \emph{Proceedings of the Third International Workshop on Paraphrasing ({IWP}2005)}.

\bibitem[{Gallegos et~al.(2023)Gallegos, Rossi, Barrow, Tanjim, Kim, Dernoncourt, Yu, Zhang, and Ahmed}]{gallegos2023bias}
Isabel~O Gallegos, Ryan~A Rossi, Joe Barrow, Md~Mehrab Tanjim, Sungchul Kim, Franck Dernoncourt, Tong Yu, Ruiyi Zhang, and Nesreen~K Ahmed. 2023.
\newblock Bias and fairness in large language models: A survey.
\newblock \emph{arXiv preprint arXiv:2309.00770}.

\bibitem[{Hinton and Roweis(2002)}]{hinton2002stochastic}
Geoffrey~E Hinton and Sam Roweis. 2002.
\newblock Stochastic neighbor embedding.
\newblock \emph{Advances in neural information processing systems}, 15.

\bibitem[{Hu et~al.(2016)Hu, Aggarwal, Ma, and Huai}]{hu2016embedding}
Renjun Hu, Charu~C Aggarwal, Shuai Ma, and Jinpeng Huai. 2016.
\newblock An embedding approach to anomaly detection.
\newblock In \emph{2016 IEEE 32nd International Conference on Data Engineering (ICDE)}, pages 385--396. IEEE.

\bibitem[{Huang et~al.(2020)Huang, Sharma, Sun, Xia, Zhang, Pronin, Padmanabhan, Ottaviano, and Yang}]{huang2020embedding}
Jui-Ting Huang, Ashish Sharma, Shuying Sun, Li~Xia, David Zhang, Philip Pronin, Janani Padmanabhan, Giuseppe Ottaviano, and Linjun Yang. 2020.
\newblock Embedding-based retrieval in facebook search.
\newblock In \emph{Proceedings of the 26th ACM SIGKDD International Conference on Knowledge Discovery \& Data Mining}, pages 2553--2561.

\bibitem[{John et~al.(2023)John, Shobayo, and Ogunleye}]{john2023exploration}
Jeen~Mary John, Olamilekan Shobayo, and Bayode Ogunleye. 2023.
\newblock An exploration of clustering algorithms for customer segmentation in the uk retail market.
\newblock \emph{Analytics}, 2(4):809--823.

\bibitem[{Kaneko and Bollegala(2021)}]{kaneko2021debiasing}
Masahiro Kaneko and Danushka Bollegala. 2021.
\newblock Debiasing pre-trained contextualised embeddings.
\newblock \emph{arXiv preprint arXiv:2101.09523}.

\bibitem[{Krieg et~al.(2023)Krieg, Parada-Cabaleiro, Medicus, Lesota, Schedl, and Rekabsaz}]{krieg2023grep}
Klara Krieg, Emilia Parada-Cabaleiro, Gertraud Medicus, Oleg Lesota, Markus Schedl, and Navid Rekabsaz. 2023.
\newblock Grep-biasir: A dataset for investigating gender representation bias in information retrieval results.
\newblock In \emph{Proceedings of the 2023 Conference on Human Information Interaction and Retrieval}, pages 444--448.

\bibitem[{Leteno et~al.(2023)Leteno, Gourru, Laclau, Emonet, and Gravier}]{leteno2023fair}
Thibaud Leteno, Antoine Gourru, Charlotte Laclau, R{\'e}mi Emonet, and Christophe Gravier. 2023.
\newblock Fair text classification with wasserstein independence.
\newblock \emph{arXiv preprint arXiv:2311.12689}.

\bibitem[{Levesque et~al.(2012)Levesque, Davis, and Morgenstern}]{levesque2012winograd}
Hector Levesque, Ernest Davis, and Leora Morgenstern. 2012.
\newblock The winograd schema challenge.
\newblock In \emph{Thirteenth international conference on the principles of knowledge representation and reasoning}.

\bibitem[{Liang et~al.(2020)Liang, Li, Zheng, Lim, Salakhutdinov, and Morency}]{liang2020towards}
Paul~Pu Liang, Irene~Mengze Li, Emily Zheng, Yao~Chong Lim, Ruslan Salakhutdinov, and Louis-Philippe Morency. 2020.
\newblock Towards debiasing sentence representations.
\newblock \emph{arXiv preprint arXiv:2007.08100}.

\bibitem[{Liu et~al.(2019)Liu, Ott, Goyal, Du, Joshi, Chen, Levy, Lewis, Zettlemoyer, and Stoyanov}]{liu2019roberta}
Yinhan Liu, Myle Ott, Naman Goyal, Jingfei Du, Mandar Joshi, Danqi Chen, Omer Levy, Mike Lewis, Luke Zettlemoyer, and Veselin Stoyanov. 2019.
\newblock Roberta: A robustly optimized bert pretraining approach.
\newblock \emph{arXiv preprint arXiv:1907.11692}.

\bibitem[{Mary et~al.(2019)Mary, Calauzenes, and El~Karoui}]{mary2019fairness}
J{\'e}r{\'e}mie Mary, Cl{\'e}ment Calauzenes, and Noureddine El~Karoui. 2019.
\newblock Fairness-aware learning for continuous attributes and treatments.
\newblock In \emph{International Conference on Machine Learning}, pages 4382--4391. PMLR.

\bibitem[{May et~al.(2019)May, Wang, Bordia, Bowman, and Rudinger}]{may2019measuring}
Chandler May, Alex Wang, Shikha Bordia, Samuel~R Bowman, and Rachel Rudinger. 2019.
\newblock On measuring social biases in sentence encoders.
\newblock \emph{arXiv preprint arXiv:1903.10561}.

\bibitem[{Mehrabi et~al.(2021)Mehrabi, Morstatter, Saxena, Lerman, and Galstyan}]{mehrabi2021survey}
Ninareh Mehrabi, Fred Morstatter, Nripsuta Saxena, Kristina Lerman, and Aram Galstyan. 2021.
\newblock A survey on bias and fairness in machine learning.
\newblock \emph{ACM computing surveys (CSUR)}, 54(6):1--35.

\bibitem[{Meng et~al.(2022)Meng, Huang, Zhang, and Han}]{meng2022generating}
Yu~Meng, Jiaxin Huang, Yu~Zhang, and Jiawei Han. 2022.
\newblock Generating training data with language models: Towards zero-shot language understanding.
\newblock \emph{Advances in Neural Information Processing Systems}, 35:462--477.

\bibitem[{Nadeem et~al.(2020)Nadeem, Bethke, and Reddy}]{nadeem2020stereoset}
Moin Nadeem, Anna Bethke, and Siva Reddy. 2020.
\newblock Stereoset: Measuring stereotypical bias in pretrained language models.
\newblock \emph{arXiv preprint arXiv:2004.09456}.

\bibitem[{Nangia et~al.(2020)Nangia, Vania, Bhalerao, and Bowman}]{nangia2020crows}
Nikita Nangia, Clara Vania, Rasika Bhalerao, and Samuel~R Bowman. 2020.
\newblock Crows-pairs: A challenge dataset for measuring social biases in masked language models.
\newblock \emph{arXiv preprint arXiv:2010.00133}.

\bibitem[{Nissim et~al.(2020)Nissim, van Noord, and van~der Goot}]{nissim2020fair}
Malvina Nissim, Rik van Noord, and Rob van~der Goot. 2020.
\newblock Fair is better than sensational: Man is to doctor as woman is to doctor.
\newblock \emph{Computational Linguistics}, 46(2):487--497.

\bibitem[{Packer et~al.(2018)Packer, Halpern, Guajardo-Cspedes, and Mitchell}]{packer2018text}
Ben Packer, Yoni Halpern, Mario Guajardo-Cspedes, and Margaret Mitchell. 2018.
\newblock Text embedding models contain bias. here’s why that matters.
\newblock \emph{Google Developers}.

\bibitem[{Palangi et~al.(2016)Palangi, Deng, Shen, Gao, He, Chen, Song, and Ward}]{palangi2016deep}
Hamid Palangi, Li~Deng, Yelong Shen, Jianfeng Gao, Xiaodong He, Jianshu Chen, Xinying Song, and Rabab Ward. 2016.
\newblock Deep sentence embedding using long short-term memory networks: Analysis and application to information retrieval.
\newblock \emph{IEEE/ACM Transactions on Audio, Speech, and Language Processing}, 24(4):694--707.

\bibitem[{Peltonen et~al.(2023)Peltonen, Xu, Nummenmaa, and Nummenmaa}]{peltonen2023fair}
Jaakko Peltonen, Wen Xu, Timo Nummenmaa, and Jyrki Nummenmaa. 2023.
\newblock Fair neighbor embedding.
\newblock In \emph{International Conference on Machine Learning}, pages 27564--27584. PMLR.

\bibitem[{Pogodin et~al.(2022)Pogodin, Deka, Li, Sutherland, Veitch, and Gretton}]{pogodin2022efficient}
Roman Pogodin, Namrata Deka, Yazhe Li, Danica~J Sutherland, Victor Veitch, and Arthur Gretton. 2022.
\newblock Efficient conditionally invariant representation learning.
\newblock \emph{arXiv preprint arXiv:2212.08645}.

\bibitem[{Rabelo et~al.(2022)Rabelo, Goebel, Kim, Kano, Yoshioka, and Satoh}]{rabelo2022overview}
Juliano Rabelo, Randy Goebel, Mi-Young Kim, Yoshinobu Kano, Masaharu Yoshioka, and Ken Satoh. 2022.
\newblock Overview and discussion of the competition on legal information extraction/entailment (coliee) 2021.
\newblock \emph{The Review of Socionetwork Strategies}, 16(1):111--133.

\bibitem[{Radford et~al.(2021)Radford, Kim, Hallacy, Ramesh, Goh, Agarwal, Sastry, Askell, Mishkin, Clark et~al.}]{radford2021learning}
Alec Radford, Jong~Wook Kim, Chris Hallacy, Aditya Ramesh, Gabriel Goh, Sandhini Agarwal, Girish Sastry, Amanda Askell, Pamela Mishkin, Jack Clark, et~al. 2021.
\newblock Learning transferable visual models from natural language supervision.
\newblock In \emph{International conference on machine learning}, pages 8748--8763. PMLR.

\bibitem[{Schick and Sch{\"u}tze(2021)}]{schick2021generating}
Timo Schick and Hinrich Sch{\"u}tze. 2021.
\newblock Generating datasets with pretrained language models.
\newblock \emph{arXiv preprint arXiv:2104.07540}.

\bibitem[{Shao et~al.(2023)Shao, Gong, Shen, Huang, Duan, and Chen}]{shao2023synthetic}
Zhihong Shao, Yeyun Gong, Yelong Shen, Minlie Huang, Nan Duan, and Weizhu Chen. 2023.
\newblock Synthetic prompting: Generating chain-of-thought demonstrations for large language models.
\newblock \emph{arXiv preprint arXiv:2302.00618}.

\bibitem[{Shen et~al.(2021)Shen, Han, Cohn, Baldwin, and Frermann}]{liu2021contrastive}
Aili Shen, Xudong Han, Trevor Cohn, Timothy Baldwin, and Lea Frermann. 2021.
\newblock \href {http://arxiv.org/abs/2109.10645} {Contrastive learning for fair representations}.
\newblock \emph{CoRR}, abs/2109.10645.

\bibitem[{Shen et~al.(2022)Shen, Han, Cohn, Baldwin, and Frermann}]{shen-etal-2022-representational}
Aili Shen, Xudong Han, Trevor Cohn, Timothy Baldwin, and Lea Frermann. 2022.
\newblock \href {https://aclanthology.org/2022.findings-aacl.8} {Does representational fairness imply empirical fairness?}
\newblock In \emph{Findings of the Association for Computational Linguistics: AACL-IJCNLP 2022}, pages 81--95, Online only. Association for Computational Linguistics.

\bibitem[{Socher et~al.(2013)Socher, Perelygin, Wu, Chuang, Manning, Ng, and Potts}]{socher-etal-2013-recursive}
Richard Socher, Alex Perelygin, Jean Wu, Jason Chuang, Christopher~D. Manning, Andrew Ng, and Christopher Potts. 2013.
\newblock \href {https://aclanthology.org/D13-1170} {Recursive deep models for semantic compositionality over a sentiment treebank}.
\newblock In \emph{Proceedings of the 2013 Conference on Empirical Methods in Natural Language Processing}, pages 1631--1642, Seattle, Washington, USA. Association for Computational Linguistics.

\bibitem[{Team et~al.(2023)Team, Anil, Borgeaud, Wu, Alayrac, Yu, Soricut, Schalkwyk, Dai, Hauth et~al.}]{team2023gemini}
Gemini Team, Rohan Anil, Sebastian Borgeaud, Yonghui Wu, Jean-Baptiste Alayrac, Jiahui Yu, Radu Soricut, Johan Schalkwyk, Andrew~M Dai, Anja Hauth, et~al. 2023.
\newblock Gemini: a family of highly capable multimodal models.
\newblock \emph{arXiv preprint arXiv:2312.11805}.

\bibitem[{Tiedemann(2012)}]{tiedemann2012parallel}
J{\"o}rg Tiedemann. 2012.
\newblock Parallel data, tools and interfaces in opus.
\newblock In \emph{Lrec}, volume 2012, pages 2214--2218. Citeseer.

\bibitem[{Wang et~al.(2023)Wang, Cheng, and Henao}]{wang2023toward}
Rui Wang, Pengyu Cheng, and Ricardo Henao. 2023.
\newblock Toward fairness in text generation via mutual information minimization based on importance sampling.
\newblock In \emph{International Conference on Artificial Intelligence and Statistics}, pages 4473--4485. PMLR.

\bibitem[{Wang et~al.(2019)Wang, Ma, Chen, Luo, Yi, and Bailey}]{wang2019symmetric}
Yisen Wang, Xingjun Ma, Zaiyi Chen, Yuan Luo, Jinfeng Yi, and James Bailey. 2019.
\newblock Symmetric cross entropy for robust learning with noisy labels.
\newblock In \emph{Proceedings of the IEEE/CVF international conference on computer vision}, pages 322--330.

\bibitem[{Wu et~al.(2024)Wu, Waheed, Zhang, Abdul-Mageed, and Aji}]{wu-etal-2024-lamini}
Minghao Wu, Abdul Waheed, Chiyu Zhang, Muhammad Abdul-Mageed, and Alham~Fikri Aji. 2024.
\newblock \href {https://aclanthology.org/2024.eacl-long.57} {{L}a{M}ini-{LM}: A diverse herd of distilled models from large-scale instructions}.
\newblock In \emph{Proceedings of the 18th Conference of the European Chapter of the Association for Computational Linguistics (Volume 1: Long Papers)}, pages 944--964, St. Julian{'}s, Malta. Association for Computational Linguistics.

\bibitem[{Yang et~al.(2023)Yang, Yu, Fung, Li, and Ji}]{yang2023adept}
Ke~Yang, Charles Yu, Yi~R Fung, Manling Li, and Heng Ji. 2023.
\newblock Adept: A debiasing prompt framework.
\newblock In \emph{Proceedings of the AAAI Conference on Artificial Intelligence}, volume~37, pages 10780--10788.

\bibitem[{Ye et~al.(2022)Ye, Gao, Li, Xu, Feng, Wu, Yu, and Kong}]{ye2022zerogen}
Jiacheng Ye, Jiahui Gao, Qintong Li, Hang Xu, Jiangtao Feng, Zhiyong Wu, Tao Yu, and Lingpeng Kong. 2022.
\newblock Zerogen: Efficient zero-shot learning via dataset generation.
\newblock \emph{arXiv preprint arXiv:2202.07922}.

\bibitem[{Yin et~al.(2019)Yin, Hay, and Roth}]{yin2019benchmarking}
Wenpeng Yin, Jamaal Hay, and Dan Roth. 2019.
\newblock Benchmarking zero-shot text classification: Datasets, evaluation and entailment approach.
\newblock \emph{arXiv preprint arXiv:1909.00161}.

\bibitem[{Yu et~al.(2023)Yu, Zhuang, Zhang, Meng, Ratner, Krishna, Shen, and Zhang}]{yu2023large}
Yue Yu, Yuchen Zhuang, Jieyu Zhang, Yu~Meng, Alexander Ratner, Ranjay Krishna, Jiaming Shen, and Chao Zhang. 2023.
\newblock Large language model as attributed training data generator: A tale of diversity and bias.
\newblock \emph{arXiv preprint arXiv:2306.15895}.

\bibitem[{Zerveas et~al.(2022)Zerveas, Rekabsaz, Cohen, and Eickhoff}]{zerveas2022mitigating}
George Zerveas, Navid Rekabsaz, Daniel Cohen, and Carsten Eickhoff. 2022.
\newblock Mitigating bias in search results through contextual document reranking and neutrality regularization.
\newblock In \emph{Proceedings of the 45th International ACM SIGIR Conference on Research and Development in Information Retrieval}, pages 2532--2538.

\bibitem[{Zhang et~al.(2024)Zhang, Cai, Li, Wu, Hou, and Abdul-Mageed}]{zhang-etal-2024-distilling}
Chiyu Zhang, Honglong Cai, Yuezhang Li, Yuexin Wu, Le~Hou, and Muhammad Abdul-Mageed. 2024.
\newblock \href {https://aclanthology.org/2024.naacl-srw.21} {Distilling text style transfer with self-explanation from {LLM}s}.
\newblock In \emph{Proceedings of the 2024 Conference of the North American Chapter of the Association for Computational Linguistics: Human Language Technologies (Volume 4: Student Research Workshop)}, pages 200--211, Mexico City, Mexico. Association for Computational Linguistics.

\bibitem[{Zhang et~al.(2016)Zhang, Yuan, Lian, Xie, and Ma}]{zhang2016collaborative}
Fuzheng Zhang, Nicholas~Jing Yuan, Defu Lian, Xing Xie, and Wei-Ying Ma. 2016.
\newblock Collaborative knowledge base embedding for recommender systems.
\newblock In \emph{Proceedings of the 22nd ACM SIGKDD international conference on knowledge discovery and data mining}, pages 353--362.

\bibitem[{Zhao and Gordon(2022)}]{zhao2022inherent}
Han Zhao and Geoffrey~J Gordon. 2022.
\newblock Inherent tradeoffs in learning fair representations.
\newblock \emph{The Journal of Machine Learning Research}, 23(1):2527--2552.

\bibitem[{Zhao et~al.(2018)Zhao, Wang, Yatskar, Ordonez, and Chang}]{zhao2018gender}
Jieyu Zhao, Tianlu Wang, Mark Yatskar, Vicente Ordonez, and Kai-Wei Chang. 2018.
\newblock Gender bias in coreference resolution: Evaluation and debiasing methods.
\newblock \emph{arXiv preprint arXiv:1804.06876}.

\bibitem[{Zliobaite(2015)}]{zliobaite2015relation}
Indre Zliobaite. 2015.
\newblock On the relation between accuracy and fairness in binary classification.
\newblock \emph{arXiv preprint arXiv:1505.05723}.

\end{thebibliography}
\clearpage
\appendix
\onecolumn

\section{Algorithm Details}
\subsection{Notation}
\begin{table}[htbp]
\begin{center}
\resizebox{0.6\columnwidth}{!}{%
\begin{tabular}{r c p{9cm} }
\toprule\hline
\multicolumn{3}{c}{\underline{Basic Variables}} \\
$L$ & $\triangleq$ & loss function\\ 
$f$,$f^{ori}$ & $\triangleq$ & finetuned and original text embedding model.   \\
$h$ & $\triangleq$ & Large language model.\\
$\theta_{p}$ & $\triangleq$ & Few-shot prompts that used to empower a LLM.\\
$A$, $a_i$ & $\triangleq$ & Sensitive attribute set and $i$-th sensitive attribute.\\ 
$S^{a_i}$,$S^n$ & $\triangleq$ & Text that relate to sensitive attribute $a_i$ and neutral text.\\
$C$,$C'$ & $\triangleq$ & Content variable and predicted content.  \\
$X^{a_i}$, $X^n$& $\triangleq$ & words from group $a_i$ and neutral words in a text. \\
$V^{a_i}$ & $\triangleq$ & words list that contains all collected words related to attribute $a_i$. \\
\hline\bottomrule
\end{tabular}
}
\caption{ Main notations used in this paper.}\label{tab:Notation}
\end{center}
\end{table}
\subsection{
The significance of text embedding fairness and its distinction from subsequent task fairness}\label{sec:diff_s_i}
Recently~\cite{liu2021contrastive, shen-etal-2022-representational} apply contrastive learning losses to mitigate biases in language representations for text classification and ~\cite{leteno2023fair,shen-etal-2022-representational} find a representational fairness and subsequent task group fairness are not, or only partially, correlated. However, subsequent tasks and text embedding fairness represent two distinct areas that are both important and need to be distinguish:

\noindent\textbf{The importance of embedding fairness:} Recent efforts, as highlighted in the introduction of our paper, emphasize the significance of text embedding fairness. The fairness of embeddings is essential due to their widespread application across various systems. For instance, Search Engine~\cite{huang2020embedding}, preprocess all content—including documents, videos, and audios—into embeddings to save on storage. When a search query is submitted, it is converted into an embedding to retrieve the most relevant results, especially during the recall phase, where embedding similarity is used to filter through numerous documents to find pertinent ones. Moreover, embeddings are directly used in other applications such as zero-shot classification~\cite{yin2019benchmarking,radford2021learning}, clustering~\cite{john2023exploration}, and Anomaly Detection~\cite{hu2016embedding}, among others. Given the critical role that embeddings play in these and additional applications, addressing fairness issues within the embeddings themselves is undeniably crucial.

\noindent\textbf{Difference between embedding fairness and subsequent task group fairness:} This paper focuses on the intrinsic fairness of text embeddings. However, the group fairness of subsequent tasks extends beyond this, incorporating additional modules that take embeddings as input for predictions, which are influenced by other sources of bias. For instance, in a medical report dataset where only females are depicted as having a cold, even if the embedding captures information about gender equally (as defined in Definition~\ref{def:fairness}), subsequent modules in the system might still incorrectly associate women with having colds. As a result, it is important to distinguish the difference between the fairness of subsequent tasks and the intrinsic fairness of embeddings.

\noindent\textbf{What we explored and can explore in the future:} In this paper, we focus on text embedding fairness and studied its influence on information retrieval tasks, as shown in Table~\ref{tab:perf} and  Table~\ref{tab:perf_v2} in our paper. Creating fair text embeddings directly improves the fairness of information retrieval. While group fairness of subsequent tasks falls outside the scope of this paper, exploring the relationship between embedding fairness and group fairness in future work could be valuable. This exploration would involve selecting an appropriate metric~\cite{mehrabi2021survey} for representation fairness and disentangle the fairness of subsequent task modules and embedding intrinsic fairness.

\noindent Considering the widespread use of embeddings, differences between group fairness and embedding fairness, we believe the fairness of text embeddings is indeed an important research topic in itself.
\subsection{Dataset Details}\label{sec:datadetail}
We generated training data using the News-Commentary-v15 corpus~\cite{tiedemann2012parallel} focusing on gender bias. By employing Gemini and ChatGPT for data augmentation, we obtained datasets comprising 43,221 and 42,930 sample pairs, respectively. Each pair contains texts with identical content from male, female, and neutral perspectives. We use last 1000 data as validation set and the remaining data as training set.
\\
For the bias evaluation dataset, we provide detailed statistics in Table~\ref{tab:sta}. Our augmented dataset sets a new benchmark, featuring an extensive dataset size that enhances the robustness and comprehensiveness of bias assessment.
\begin{table}[h!]
\centering
\begin{tabular}{lll}
\hline\hline
\textbf{Evaluation Data}                  & \textbf{Level}  & \textbf{Data Size} \\ \hline
Sentence Encoder Association Test (SEAT)     & Text           & 5172               \\ 
CrowS-Pairs & Text           & 1508               \\ 
StereoType Analysis                   & Text                  & 8497               \\
Gender-Bias-IR                       & Query-Doc      & 236                \\ 
CCD-GPT (ours)                     & Text      & 42,930             \\
CCD-Gemini (ours)                        & Text      &  43,221                 \\ \hline\hline
\end{tabular}
\caption{Dataset Statistics on various bias evaluation benchmarks.}
\label{tab:sta}
\end{table}

\subsection{Data Augmentation Prompts}
The prompt template can be found in Figure~\ref{fig:fair}. To provide a clearer demonstration, we also list the examples we used. Notably, to save computational costs, we have shortened the examples and merged the selected 10 examples into 8, as shown in the Table~\ref{tab:task_template}.
\subsection{Ommited Proofs}\label{sec:proof}
In this section, we give a detailed proof of Theorem~\ref{the:theory}. 
\begin{proof} 
Firstly, we establish the conditional independence \( A \perp C' \mid C \) for any \( a_i, a_j \in A \):
\begin{align}\label{eq:cond}
    P(C' \mid A = a_i, C) = P(C' \mid A = a_j, C)
\end{align}
where \( C' \) represents the content embedding. Assuming equal probabilities for different sensitive attributes \( P(a_1 \mid C) = \cdots = P(a_A \mid C) \), we can rewrite Eq.~\eqref{eq:cond} as:
\begin{align}
P(C' \mid A = a_i, C)P(a_i \mid C) &= P(C' \mid A = a_j, C)P(a_j \mid C) \nonumber\\
     P(C', a_i \mid C) &= P(C', a_j \mid C)
\end{align}
According to Section~\ref{sec:prob}, \( f(S_C^{a_i}) \) encodes both content and sensitive information, allowing us to obtain:
\begin{align}\label{eq:eqp}
      P(f(S_C^{a_i}) \mid C) &= P(f(S_C^{a_j}) \mid C)
\end{align}
Because a fair and well-trained embedding model \( f \) can effectively extract the content \( C \) from the neutral text \( S^n_C \) without introducing bias, we can approximate Eq.~\eqref{eq:eqp} as:
\begin{align}\label{eq:eqp2}
      P(f(S_C^{a_i}) \mid f(S_C^{n})) &= P(f(S_C^{a_j}) \mid f(S_C^{n}))
\end{align}
Following~\cite{hinton2002stochastic,yang2023adept}, the conditional probability \( P(f(S_C^{a_i}) \mid f(S_C^{n})) \) can be represented as the similarity between \( S_C^{a_i} \) and \( f(S_C^{n}) \), and can be modeled using a Gaussian distribution. We thus measuring \( P(f(S_C^{a_i}) \mid f(S_C^{n})) \) by calculating:
\begin{align}\label{eq:pp}
   P (f(S_C^{a_i})  \mid f(S_C^{n})) = \frac{\exp\left(- \frac{\lVert f(S_C^{a_i}) - f(S_C^n) \rVert^2}{2\rho^2} \right)}
{\sum_{a_i \in A} \exp\left(- \frac{\lVert f(S_C^{a_i}) - f(S_C^n) \rVert^2}{2\rho^2} \right)}
\end{align}
where $\rho$ controls falloff of the $P$ with respect to distance and is set by hand. Eq.~\eqref{eq:pp} can be interpreted as follows: (1) Consider setting a Gaussian distribution with a covariance matrix equal to $\rho$ times the identity matrix at the embedding of a neutral text $S_C$ (with content $C$), which is denoted as $f(S_C^n)$. Then, a text with the same content but containing sensitive information $a_i$ appears in the distribution with a probability proportional to $\exp\left(- \frac{\lVert f(S_C^{a_i}) - f(S_C^n) \rVert^2}{2\rho^2}\right)$, represented as the numerator. (2) The denominator aggregates the aforementioned probabilities across all sensitive groups $a_i \in A$ and serves as the normalization factor.
Then we combine Eq.~\eqref{eq:eqp2} and Eq.~\eqref{eq:pp} and obtain:
\begin{align}
    \frac{\exp\left(- \frac{\lVert f(S_C^{a_i}) - f(S_C^n) \rVert^2}{2\rho^2} \right)}
{\sum_{a_i \in A} \exp\left(- \frac{\lVert f(S_C^{a_i}) - f(S_C^n) \rVert^2}{2\rho^2} \right)} &= \frac{\exp\left(- \frac{\lVert f(S_C^{a_j}) - f(S_C^n) \rVert^2}{2\rho^2} \right)}
{\sum_{a_j \in A} \exp\left(- \frac{\lVert f(S_C^{a_j}) - f(S_C^n) \rVert^2}{2\rho^2} \right)} \nonumber\\
\exp\left(- \frac{\lVert f(S_C^{a_i}) - f(S_C^n) \rVert^2}{2\rho^2} \right) &= \exp\left(- \frac{\lVert f(S_C^{a_i}) - f(S_C^n) \rVert^2}{2\rho^2} \right) \nonumber\\
\lVert f(S_C^{a_i}) - f(S_C^n) \rVert^2 &= \lVert f(S_C^{a_j}) - f(S_C^n) \rVert^2
\end{align}
Thus we obtain the Theorem~\ref{the:theory}. As a result, achieving conditional independence between sensitive attributes and content embeddings is equivalent to achieving content-conditioned equal distance. 
\end{proof} 

 \begin{table*}[ht!]
    \centering
    \resizebox{\textwidth}{!}{%
    \begin{tabular}{p{2cm}|p{3.5cm}|p{3.5cm}|p{3.5cm}|p{3.5cm}}
        \hline
         \hline
        \textbf{Example} & \textbf{Original passage} & \textbf{Neutral passage} & \textbf{Male passage} & \textbf{Female passage} \\ \hline
        Example 1 & The high popularity of the current president (Socialist Michelle Bachelet, Chile’s first female chief executive) & The high popularity of the current president (A Socialist, Chile’s first chief executive) & The high popularity of the current president (Socialist Mike Bachelet, Chile’s first male chief executive) & The current president (Socialist Michelle Bachelet, Chile’s first female chief executive) \\ \hline
        Example 2 & Rwanda has the highest female legislators in the world. & Rwanda has the highest legislators in the world. & Rwanda has the highest male legislators in the world. & Rwanda has the highest female legislators in the world. \\ \hline
        Example 3 & When a kid arrived, accompanied by a doting father, the prophet’s son. & When a kid arrived, accompanied by a doting parent, the prophet’s child. & When a kid arrived, accompanied by a doting father, the prophet’s son. & When a kid arrived, accompanied by a doting mother, the prophet’s daughter. \\ \hline
        Example 4 & wizards Hunt people, poor paternal nutrition. & People Hunt people, poor nutrition. & wizards Hunt people, poor paternal nutrition. & Witch Hunt people, poor maternal nutrition. \\ \hline
        Example 5 & Bruni’s life path become opera divo, barman and actress. & A people's life path become opera performer, bar staff and acting. & Michael's life path become opera diva, barwoman and actor. & Bruni’s life path become opera divo, barman and actress. \\ \hline
        Example 6 & Ally is marchioness, Bride for Sarkozy. & they are noble, partner of someone. & Alexandria is marquis, Groom for Sara. & Ally is marchioness, Bride for Sarkozy. \\ \hline
        Example 7 & Mike embarked on a fascinating experiment with sons. & Leader embarked on a fascinating experiment with offsprings. & Mike embarked on a fascinating experiment with sons. & Merkel embarked on a fascinating experiment with daughters. \\ \hline
        Example 8 & Orban and Tomy appointed a police as his secretary, most strong-minded male Democrat. & They appointed a police as their secretary, most strong-minded Democrat. & Orban and Tomy appointed a police as his secretary, most strong-minded male Democrat. & Olivia and Michelle appointed a police as her secretary, most strong-minded female Democrat. \\ \hline  \hline
    \end{tabular}
    }
    \caption{Task template and prompt examples for gender-neutral, male, and female passages.}
    \label{tab:task_template}
\end{table*}

\end{document}